\documentclass{article}

\usepackage{PRIMEarxiv}

\usepackage[utf8]{inputenc} 
\usepackage[T1]{fontenc}    
\usepackage{hyperref}       
\usepackage{url}            
\usepackage{booktabs}       
\usepackage{amsfonts}       
\usepackage{nicefrac}       
\usepackage{microtype}      
\usepackage{lipsum}
\usepackage{fancyhdr}       
\usepackage{graphicx}       
\usepackage{tikz}
\usepackage{multirow}
\usepackage[ruled, vlined]{algorithm2e}
\usepackage{algorithmicx}
\usepackage{CJKutf8}
\usepackage{amsmath}
\graphicspath{{media/}}     

\title{Overview of AI-Debater 2023: The Challenges of Argument Generation Tasks}

\author{
Jiayu Lin\textsuperscript{1}\thanks{~~Equal contribution.},
Guanrong Chen\textsuperscript{2}$^{*}$,
Bojun Jin\textsuperscript{2}$^{*}$,
Chenyang Li\textsuperscript{3}$^{*}$,
Shutong Jia\textsuperscript{4}$^{*}$,
Wancong Lin\textsuperscript{5}$^{*}$,
Yang Sun\textsuperscript{2},\\
{\bf Yuhang He\textsuperscript{2},
Caihua Yang\textsuperscript{2},
Jianzhu Bao\textsuperscript{2},
Jipeng Wu\textsuperscript{2},
Wen Su\textsuperscript{4},
Jinglu Chen\textsuperscript{4},
Xinyi Li\textsuperscript{4},}\\
{\bf Tianyu Chen\textsuperscript{4},
 Mingjie Han\textsuperscript{5},
Shuaiwen Du\textsuperscript{5},
Zijian Wang\textsuperscript{5},
Jiyin Li\textsuperscript{5},
Fuzhong Suo\textsuperscript{5},}\\
{\bf Hao Wang\textsuperscript{5},
Nuanchen Lin\textsuperscript{6},
Xuanjing Huang\textsuperscript{1},
Changjian Jiang\textsuperscript{1},
Ruifeng Xu\textsuperscript{2}\thanks{~~Corresponding author.},
Long Zhang\textsuperscript{3}$^{\dagger}$,}\\
{\bf Jiuxin Cao\textsuperscript{4}$^{\dagger}$,
Ting Jin\textsuperscript{5}$^{\dagger}$,
Zhongyu Wei\textsuperscript{1}$^{\dagger}$}\\
\textsuperscript{1}Fudan University\\
~\textsuperscript{2}Harbin Insitute of Technology, Shenzhen\\
~\textsuperscript{3}Zhongyuan University of Technology, ~\textsuperscript{4}Southeast University\\
~\textsuperscript{5}Hainan University, ~\textsuperscript{6}South China Agricultural University\\
\texttt{jiayulin22@m.fudan.edu.cn, \{23S051030, 200110828\}@stu.hit.edu.cn}\\
\texttt{\{lcy, zhanglong\}@zut.edu.cn, \{shutong\_jia, jx.cao\}@seu.edu.cn}\\
\texttt{\{wanconglin, jinting\}@hainanu.edu.cn, xuruifeng@hit.edu.cn}\\
\texttt{zywei@fudan.edu.cn}
}

\begin{document}
\begin{CJK}{UTF8}{gbsn}
\maketitle

\begin{abstract}
In this paper we present the results of the AI-Debater 2023 Challenge held by the Chinese Conference on Affect Computing (CCAC 2023), and introduce the related datasets. We organize two tracks to handle the argumentative generation tasks in different scenarios,
namely, Counter-Argument Generation (Track 1) and Claim-based Argument Generation (Track 2). Each track is equipped with its distinct dataset and baseline model respectively. In total, 32 competing teams register for the challenge, from which we received 11 successful submissions. In this paper, we will present the results of the challenge and a summary of the systems, highlighting commonalities and innovations among participating systems. Datasets and baseline models of the AI-Debater 2023 Challenge have been already released and can be accessed through the official website\footnote{\url{http://www.fudan-disc.com/sharedtask/AIDebater23/index.html}} of the challenge.
\end{abstract}


\section{Introduction}
Argument and debate are fundamental capabilities of human intelligence, essential for a wide range of human activities, and common to all human societies. Argumentation~\cite{besnard2008elements, o1927argumentation, van2015reasonableness} takes the human logical argumentation process as the research object, and is a research field involving logic, philosophy, language, rhetoric, computer science and education. Striving to enable models to automatically understand and generate argument texts, computational argumentation, a newly emerging research field, is obtaining increasing attention from the research community~\cite{slonim2021autonomous}. Depending on the task objectives, computational argumentation tasks can be divided into two aspects, argument mining and argument generation.

With the rapid development of modern technology, online forums like ChangeMyView allow people to freely exchange opinions on specific topics, making them suitable data sources for argument generation tasks, especially for designing artificial debaters, as online forums closely resemble real-world debates. Initial research in this field has focused on analyzing ChangeMyView data~\cite{seaman2016winning, wei2016post} to summarize the key factors of persuasive arguments.

For an extended period, the field of argument mining has been particularly active. Li et al.~\cite{li2022structure} proposed a Structure-Aware Argument Encoder (SAE) in their work, which enhances the ability to capture structural information in the analysis of scientific literature discourse by distinguishing between framing words and topic words in sentences and incorporating paragraph-level positional information. Additionally, some researchers have integrated knowledge graph structures into argument mining tasks. Yuan et al.~\cite{yuan2021leveraging} constructed a knowledge graph for external knowledge, improving the model's ability to identify interactive argument pairs. Liang et al.~\cite{liang2023hi} proposed a hierarchical argumentation graph structure and introduced a text-graph multi-modal pre-training framework.

Recently, large language models, such as OpenAI ChatGPT and GPT-4~\cite{bubeck2023sparks}, PaLM~\cite{chowdhery2023palm}, and LlaMAs~\cite{touvron2023llama, touvron2023llama2} have achieved great success and demonstrated remarkable performance in text generation tasks. Therefore, to align the field of computational argumentation with the development trend of large language models, we have organized the AI-Debater 2023 Challenge\footnote{This event is an CCAC 2023 task sponsored by Fudan University.}. This challenge focuses on generation tasks, including two tracks: counter-argument generation (Track 1) and claim-based argument generation (Track 2). In Track 1, we introduce the task of generating counter-argument based on given topic; while in Track 2, we introduce the task of generating argument based on given claim. We provide two datasets in this task, one for each track.

In total, 32 teams from over 10 colleges and corporates enter for the challenge, 11 of which successfully submit their models and obtain their model’s performance. We hope that we can prompt the computational argumentation community to align itself with mainstream text generation technologies through this challenge.

In this paper, we present a detailed description for each track and their dataset, along with technical solutions of the winning team, and discuss the possible future research directions of the task.
\section{Related Works}
\subsection{Counter-Argument Generation}
Datasets for counter-argument generation mainly establish the rebuttal relationship in the conversation using automatic methods such as citation or reply detection~\cite{ji2021discrete, hua2018neural}. Seaman et al.~\cite{seaman2016winning} proposed CMV dataset, including the citation relationship between original posts and their corresponding replies. Bolton et al.~\cite{bolton2020high} introduced Kialo, a dataset for sentence-level argument stance classification, which can also be applied to counter-argument generation task. Lin et al.~\cite{lin2023argue} introduced ArgTersely, a dataset for sentence-level counter-argument generation, this dataset is obtained by manual annotation.

Early work~\cite{hua2018neural, hua2019argument} focus on how to introduce external knowledge into the system; Alshomary et al.~\cite{alshomary2021counter} developed a system to identify weak points in arguments; Schiller et al.~\cite{schiller2021aspect} developed a controlled argument generation system, which is able to generate arguments based on given information; Alshomary et al.~\cite{alshomary2023conclusion} completed it through multi-task and multi-step reasoning. Lin et al.~\cite{lin2023argue} constructed argumentation instructions, and fine-tuned a large language model for this task.

\subsection{Claim-based Argument Generation}
Claim-based argument generation is a burgeoning field within NLP that aims to construct persuasive arguments automatically. This involves not only comprehending the topic but also aligning the generated claims with the audience's beliefs for increased effectiveness. 

Alshomary et al.~\cite{alshomary2021belief} address the challenge of tailoring arguments to an audience's beliefs by generating claims that are both topic-relevant and belief-aligned. Hu et al.~\cite{hu2023americano} propose AMERICANO framework and innovate argument generation through discourse-driven decomposition and agent interaction, enhancing the coherence and persuasiveness of generated arguments. Alkhawaldeh et al.~\cite{alkhawaldeh2020rl} introduces a deep learning and reinforcement learning-based approach for generating Toulmin arguments, focusing on claim and warrant components to enhance stance detection and factuality checking in NLP tasks.

\section{Task Description and Result}
In this section, we formally define the specific task, introduce the construction of the corresponding dataset, scoring metrics as well as the baseline model for each track respectively. The results of this competition can be found in the Appendix~\ref{cha_res}.
\subsection{Track 1: Counter-Argument Generation}

\paragraph{Task Formulation} We formulate our task according to Lin et al.~\cite{lin2023argue}'s setting. For a given topic $\tau$ and original argument$x$, the participating model automatically generates one sentence $y$ that refutes the original argument (referred to as a counter-argument).

\begin{equation}
    y = F_1(\tau, x)
\end{equation}

\paragraph{Data Construction} We created ArgTersely dataset for counter-argument generation task by extracting data from the ChangeMyView (CMV) debate forum and manually annotating them. The process began with data preprocessing to segment replies into sentences and remove invalid content. Annotators then selected sentences that countered the original arguments during trial annotation, which also served as training and consistency testing with reference annotations. The formal annotation phase used a cross-annotation strategy with two annotators per triplet and a third to resolve disagreements, ensuring dataset quality. During AI-Debater 2023 challenge, we used a subset of this dataset with 10,000 training and 4,000 test samples.

\paragraph{Scoring Metric} We use ROUGE-L score as the automatic evaluation metrics.
\paragraph{Baseline Model} 
We fine-tuned GPT-2~\cite{radford2019language} as a baseline model. Specifically, we concatenated the debate topic, original argument, and counter-argument into a continuous text, applied mask processing to the debate topic and original argument, and then conducted auto-regressive training targeting the counter-argument part with a cross-entropy loss function.


\subsection{Track 2: Claim-based Argument Generation}
\paragraph{Task Formulation} In this task, for the given claim $c$, the participating model automatically generates 5 independent arguments, $Z=[z_1, z_2,..., z_5]$ that fit the claim.

\begin{equation}
    z_i = F_2(c), i=1,2,...,5
\end{equation}
\paragraph{Data Construction} The dataset is derived from nearly 700 renowned Chinese debate competitions held between 2007 and 2021. Each debate match's segment and monologue text were obtained through speech-to-text transcription and subsequent manual verification. The monologue texts were chunked based on punctuation marks such as periods and question marks, and then annotators marked the argument sentences. Each argument sentence corresponds to the claim of the current debate round, resulting in pairs of claim-argument data. During AI-Debater 2023 challenge, the training set includes 33 claims with 3455 arguments, and the test set comprises 41 claims with 930 arguments.
\paragraph{Scoring Metric} We use ROUGE-L score as the automatic evaluation metrics.
\paragraph{Baseline Model} We fine-tuned Mengzi-T5-base~\cite{zhang2021mengzi} as a baseline model. Specifically, we concatenated the claim and the argument into a continuous text, applied mask processing to the debate topic, and then conducted auto-regressive training targeting the argument part with a cross-entropy loss function.


\section{Technical Approaches}
\clearpage
\subsection{Track 1: Data Augmentation and Instruction Tuning in Counter-Argument Generation}
This subsection will introduce the details of the model submitted by HITSZ-HLT team in Track 1.

\subsubsection{Analysis of the Problem}
The competition's objective is to create a model capable of automatically generating counter-arguments for a given topic and original argument. The training data set presents challenges such as duplicate topics and sources, and a skewed distribution of counter-arguments in length and frequency. 

The original arguments typically range from 30 to 200 words, averaging 108.9877 words, while counter-arguments range from 30 to 250 words, averaging 118.8507 words. The counter-argument length distribution is notably uneven, with a few excessively long sentences that can introduce noise into the training process. 

Additionally, very short sentences can impede the model's ability to learn complex logical expressions. The topic distribution is also uneven, with the majority of topics having more than forty counter-arguments, and the least having only a few.

\subsubsection{Methodology}
\begin{figure}[h]
    \centering
    \includegraphics[width=0.7\textwidth]{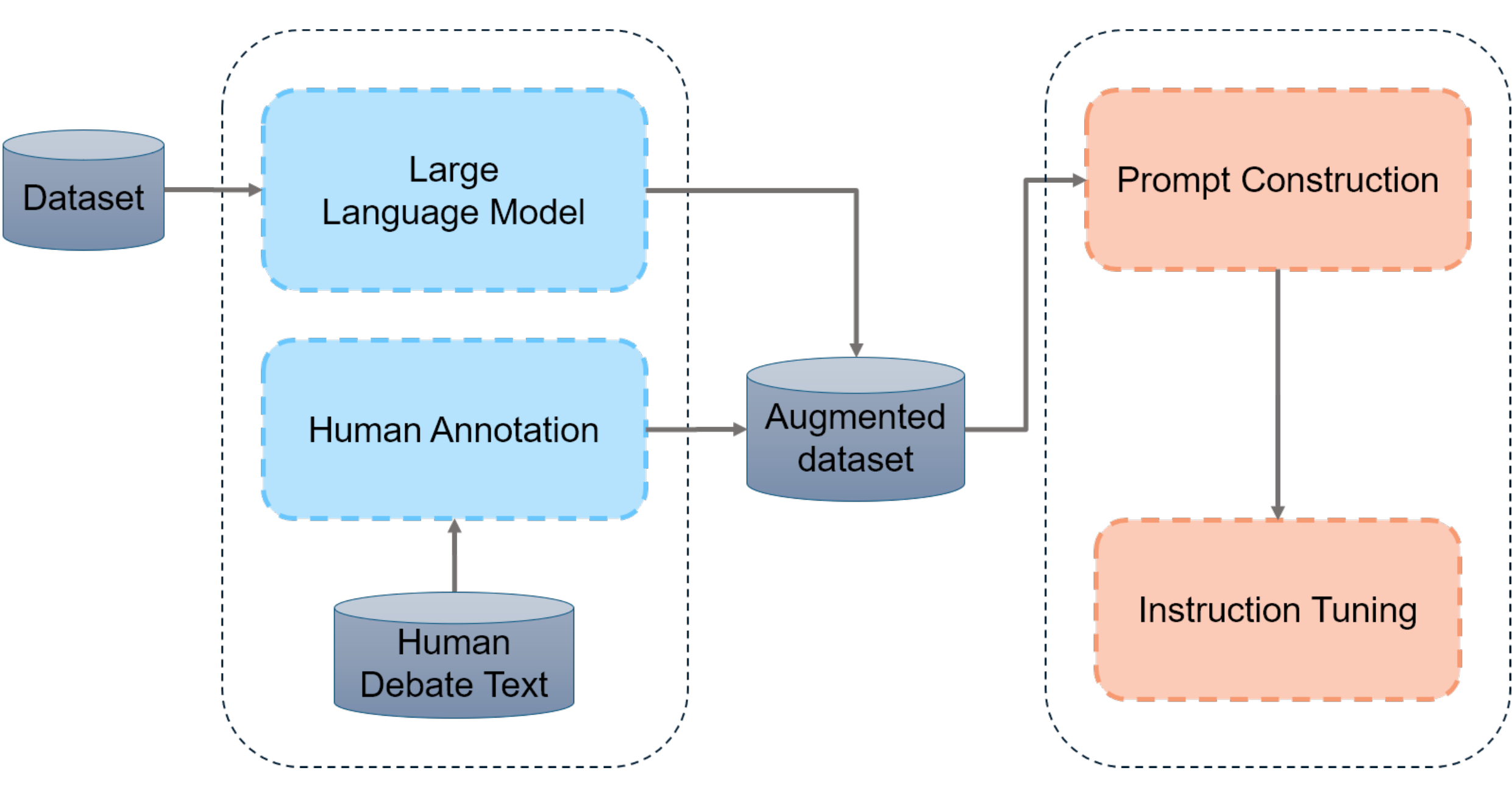}
    \caption{The overall architecture of the proposed method.}
    \label{track1hitsz1}
\end{figure}

As is shown in figure~\ref{track1hitsz1}, our methodology encompasses a two-part approach: a data augmentation module and a generative language model based on instruction tuning. The data augmentation module addresses the imbalance in the training data through two-tiered expansion. Firstly, we utilized ChatGPT~\cite{openai2022chatgpt} to generate novel counter-arguments for existing topics, adding 6171 new data points after filtering. Secondly, we incorporated human debate data from the Kialo forum, manually curating and labeling topics to add 9987 new data points and 98 new topics. We also refined the data by removing extreme lengths and low-quality text, such as profanity and non-argumentative sentences, to enhance model performance.

\begin{figure}[h]
    \centering
    \includegraphics[width=0.7\textwidth]{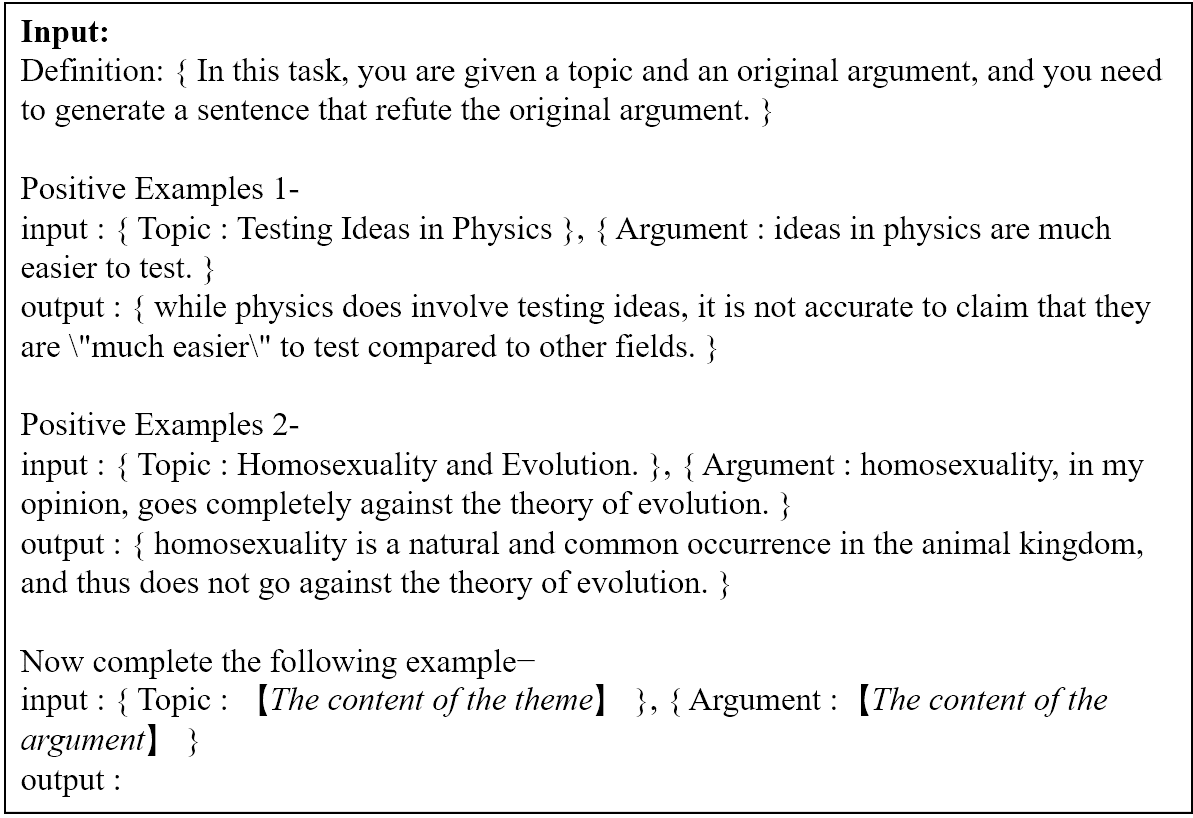}
    \caption{Instruction fine-tuning template for generating counter-arguments.}
    \label{track1hitsz2}
\end{figure}

For the generative language model, we employed instruction tuning on a pre-trained model, selecting Tk-INSTRUCT~\cite{wang2022super} over Flan-T5~\cite{chung2024scaling} for its superior performance. Tk-INSTRUCT was fine-tuned using a dataset covering 1616 diverse NLP tasks. As is shown in figure~\ref{track1hitsz2}, we crafted instruction templates tailored to the counter-argument task, consisting of a task definition, positive example demonstrations, and reasoning cases. The instruction template was designed with two positive cases, and we explored the use of connective adverbs to promote syntactic diversity in the output.

\subsubsection{Experiments}
In our experiments, the validation set was structured to reflect the original dataset's length distribution. We experimented with theme-based division but found it less effective due to the uneven distribution of themes and counter-arguments. The model implementation utilized tk-instruct-large-def-pos, a model with 770 million parameters, and applied a minimum generation length of 50 words and a maximum of 256 words to prevent the generation of overly short sentences that could reduce model performance. We employed Beam Search with three beams for decoding, balancing the decoding effect with training time efficiency. To mitigate repetitive word generation, we set no\_repeat\_ngram\_size to 2. The result is shown in table~\ref{track1hitsz_table1}.

Our model achieved a ROUGE-L score of 0.252 on the official test set, and manual inspection of the validation set output demonstrated that the model could understand and generate counter-arguments with good logical and thematic relevance. Ablation studies confirmed the positive impact of our data augmentation module, with the addition of real human debate text from Kialo proving most effective.

\begin{table}[h]
\centering
\begin{tabular}{@{}lcc@{}}
\toprule
Model & ROUGE-L \\
\midrule
w/o D    & 0.2301 \\
w/o ChatGPT & 0.2351 \\
w/o Kialo & 0.2389 \\
Our model & 0.2400 \\
\bottomrule
\end{tabular}
\caption{The impact of different data augmentation approaches.}
\label{track1hitsz_table1}
\end{table}

\clearpage
\subsection{Track 1: Pre- and Post-Processing in Counter-Argument Generation}
This subsection will introduce the details of the model submitted by huashui team in Track 1.
\subsubsection{Framework} Pre-processing and post-processing are pivotal in NLP, particularly for text generation where they encompass tokenization, template design, and decoding strategies. Despite the prevalence of pre-trained models fine-tuned for specific tasks, these methods fall short in low-resource or under-equipped settings. Our approach circumvents this by optimizing performance through strategic pre-processing and post-processing, without structural model changes. Experimental results validate the efficacy of our strategies against those reliant on extensive data or model modifications.

The overall framework of our study is as follows. Initially, the original text is transformed into an input that is more easily understood by the model through a predefined template. Subsequently, the tokenizer completes the basic word embedding and inputs it into the GPT-2 model to extract text features and predict the probability of generating words. Generally, greedy algorithms are used as the default decoding strategy in current research. However, the demand in this track is to allow the model to output multiple sentences simultaneously. Therefore, this study introduces diverse beam search~\cite{vijayakumar2016diverse} and contrastive search~\cite{su2022contrastive} into the model decoding process.

Diverse beam search~\cite{vijayakumar2016diverse} improves upon the limitations of the "single-point departure" inherent in traditional beam search strategies. It draws on the ideas of breadth-first search (BFS), exploring paths from multiple different starting points, effectively enhancing the diversity of the model's generation. Contrastive search, a new concept proposed in 2022, involves judging the text similarity matrix at each decoding moment to incorporate a similarity penalty, resulting in non-repetitive yet coherent output.

\begin{figure}[h]
    \centering
    \includegraphics[width=0.6\textwidth]{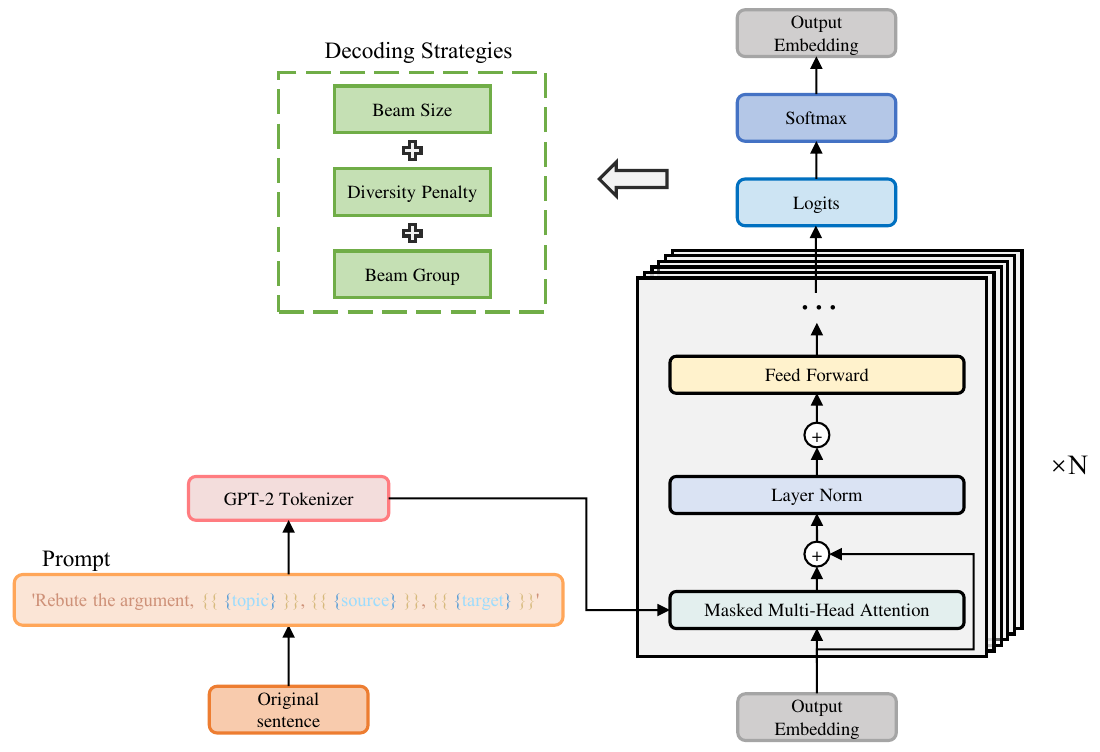}
    \caption{Overall model framework.}
    \label{track1huashui}
\end{figure}

\subsubsection{Experiments} In our experimental setup, we used the GPT-2 base version as the foundational model. For diverse beam search, the beam width was set to 5, the number of groups to 5, and the diversity penalty to 1. For contrastive search, the penalty factor was set to 0.6, and the top-k value to 5. The experimental results are as follows:

\begin{table}[htbp]
\centering
\begin{tabular}{ccc}
\hline
Decoding Strategy & ROUGE-L \\ \hline
Greedy Search (Baseline) & 0.158 \\
Contrastive Search & 0.162 \\
Diverse Beam Search & 0.172 \\
\hline
\end{tabular}
\caption{Experimental results of different decoding strategies.}
\end{table}

It can be observed that the two strategies adopted in this study have outperformed the method used by the baseline model, thereby proving the rationality of the starting point of this study.

\clearpage
\subsection{Track 1: A Diffusion Framework for Counter-Argument Generation}
This subsection will introduce the details of the model submitted by ZUT team in Track 1.
\subsubsection{Controlled Text Generation Task Formulation}
The problem addressed in this document can be defined as follows: Given control attributes (arguments, claims) $w^x$ and a target text (counter-argument) $w^y$, train a language model to output high-quality $w^y$ that aligns with the control attributes upon input $w^x$.

\begin{equation}
    p(w^{y'}|w^x) \propto p(w^{y'}) \cdot p(w^x|w^{y'})
\end{equation}

The controlled text generation task is formalized as sampling from a conditional distribution $p(w^{y'}|w^x)$, where $w^x$ represents control attributes, $p(w^{y'})$ ensuring fluency to complete the attribute control process $p(w^x|w^{y'})$.

\begin{figure}[h]
    \centering
    \includegraphics[width=0.7\textwidth]{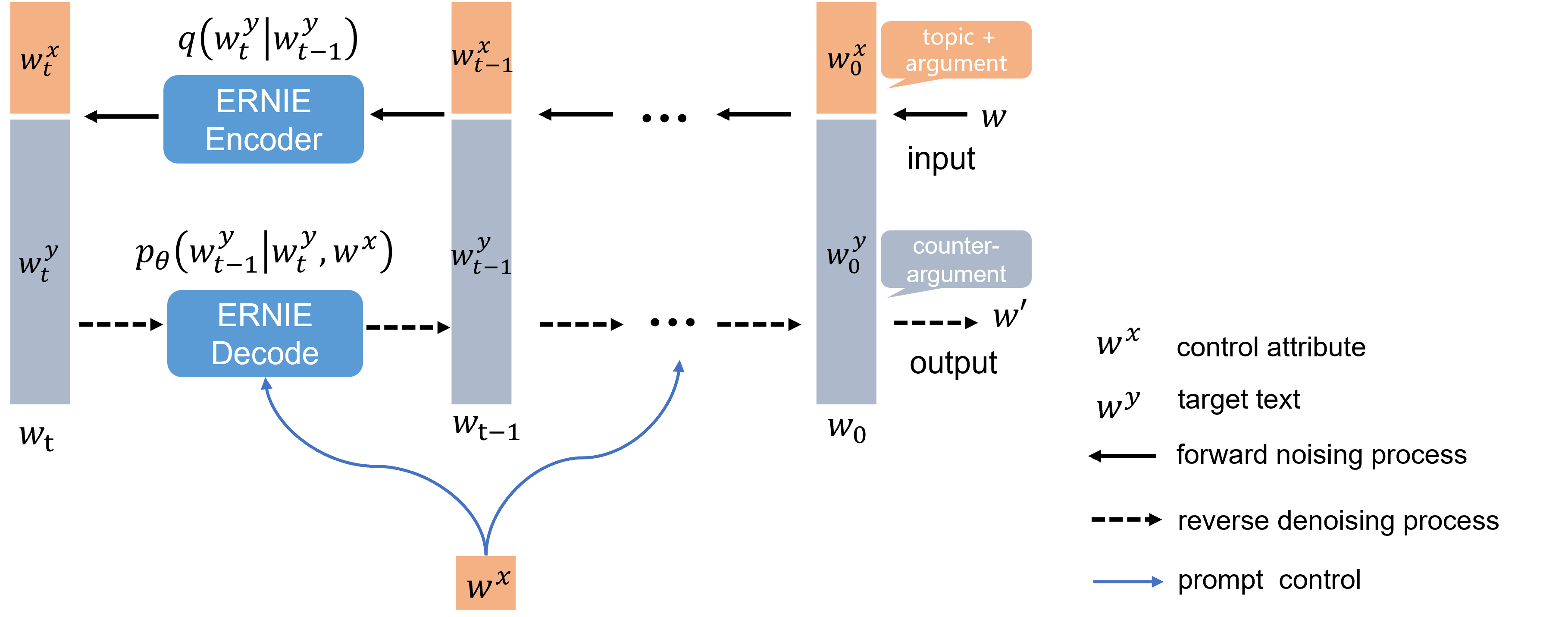}
    \caption{Sequence diffusion generation model integrating pre-trained models.}
    \label{track1zut1}
\end{figure}

\subsubsection{Sequence Diffusion Process}
Inspired by D3PM, we use a method for diffusing disorganized text by treating mask tokens as noise addition and decode tokens as noise removal during the diffusion process. The forward diffusion process involves progressively masking tokens, while the reverse diffusion process decodes the masked tokens back into text.

\begin{figure}[h]
    \centering
    \includegraphics[width=0.5\textwidth]{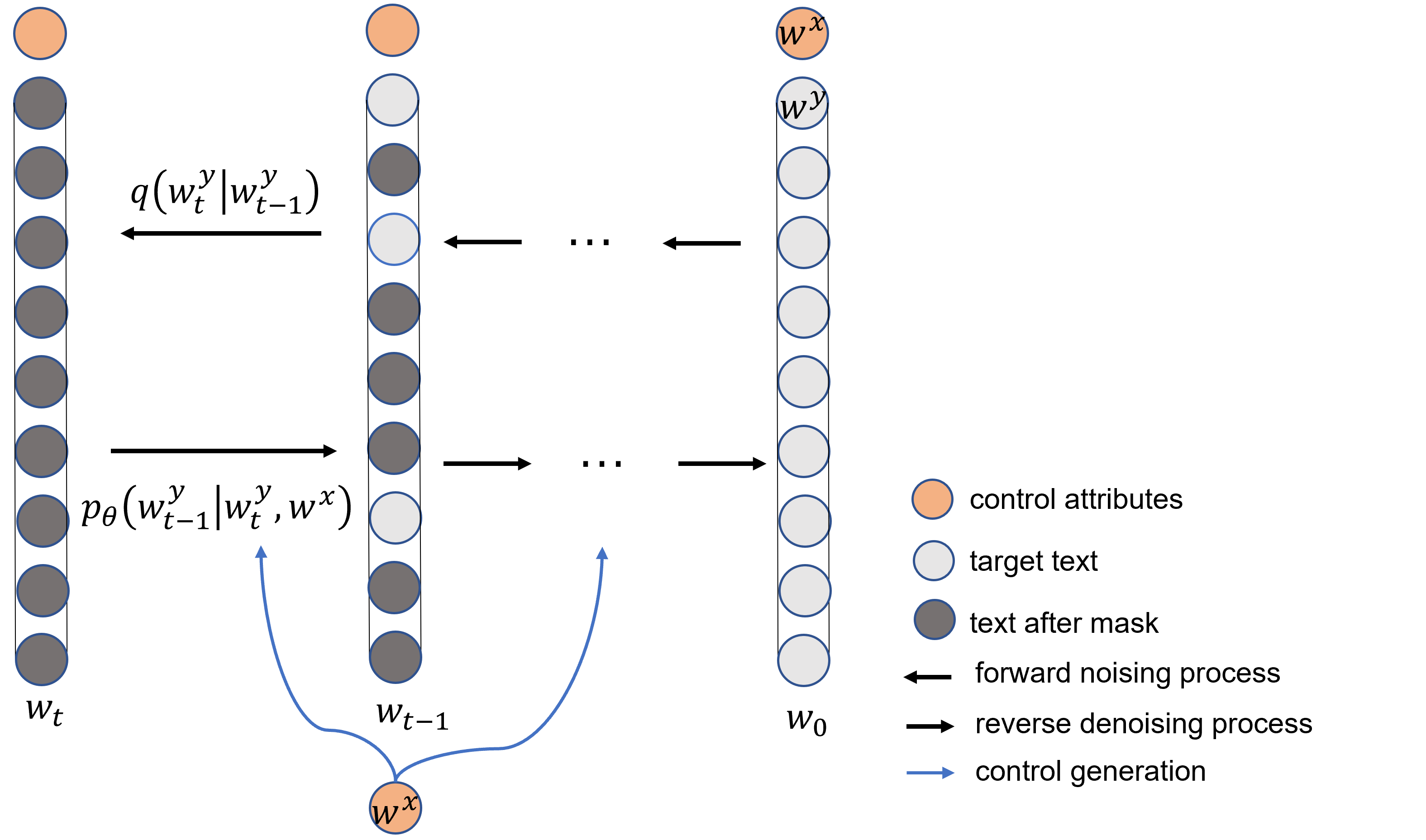}
    \caption{Sequence diffusion process.}
    \label{track1zut2}
\end{figure}

\subsubsection{Model Integration}
We introduces a combined sequence diffusion model that integrates a pretrained model (BERT) with a diffusion model. The model uses BERT's encoding and decoding capabilities in conjunction with the diffusion model's noise addition and removal processes. This integration allows for the establishment of a connection between control attributes and corresponding text within different feature spaces.

The diffusion process is guided by a posterior distribution, with specific steps outlined for optimization and regularization of fluency. The model aims to generate high-quality text with controlled attributes without the need for a separate attribute classifier, thus avoiding errors and reducing training time.

\begin{figure}[h]
    \centering
    \includegraphics[width=0.7\textwidth]{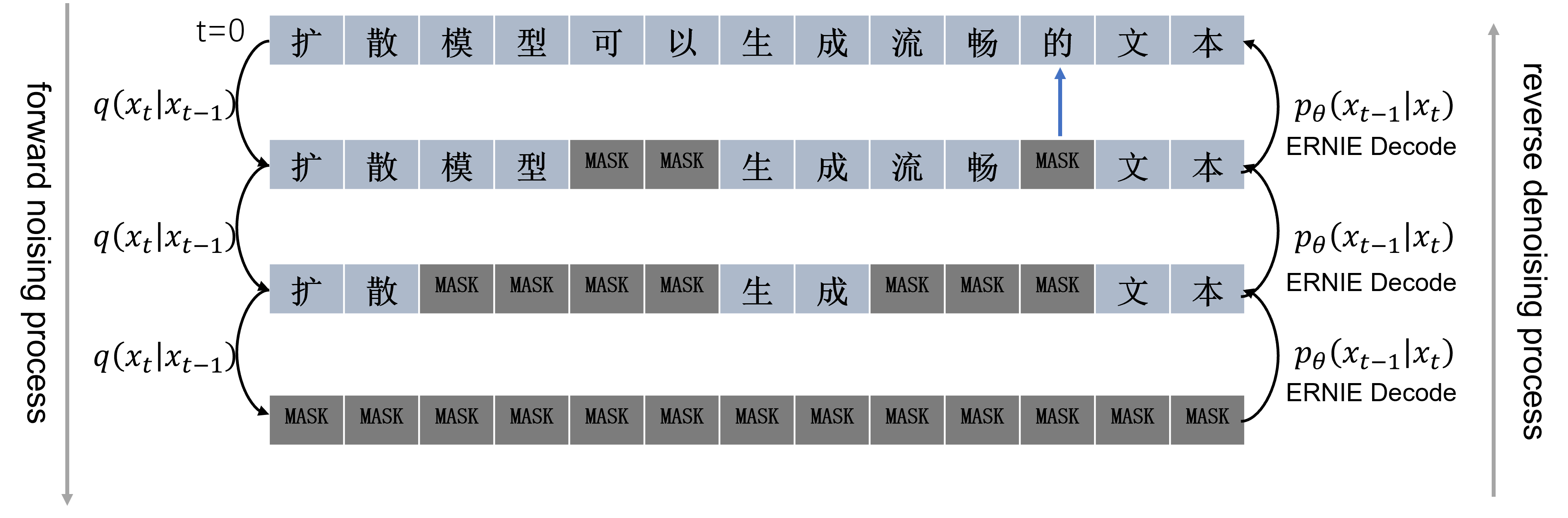}
    \caption{Integration process of diffusion model and pre-training model.}
    \label{track1zut3}
\end{figure}

\subsubsection{Experiments}
The document presents experimental results comparing the proposed model with baseline models GPT-2. The performance is measured using the ROUGE-L metric, which evaluates the quality of generated text.

\begin{table}[htbp]
\centering
\begin{tabular}{lccc}
\hline
Method & Pretrained & Step & ROUGE-L\\
\hline
Baseline & GPT-2 & 1 & 0.143 \\
\hline
\multirow{2}{*}{Our Model} & \multirow{2}{*}{BERT} & 256 & 0.159 \\
          &        & 512 & 0.188 \\
\hline
\end{tabular}
\caption{Experimental results in Track 1.}
\end{table}

\clearpage
\subsection{Track 2: Enhancing Argument Diversity for Claim-based Argument Generation}
This subsection will introduce the details of the model submitted by HITSZ-HLT team in Track 2.

\subsubsection{Task Analysis}
The competition's objective was to create an automated system capable of generating five relevant arguments for a given claim. The analysis of the competition data revealed that while claims were brief, the corresponding arguments were more extensive. The challenge lay in producing lengthy and varied texts from short inputs. With an average of 104.70 arguments per claim, the task was to efficiently utilize this wealth to generate five distinct arguments. Additionally, the dataset included some arguments that were short and lacked substance, necessitating a strategy to address these issues.

\subsubsection{Methods}
\begin{figure}[h]
    \centering
    \includegraphics[width=0.7\textwidth]{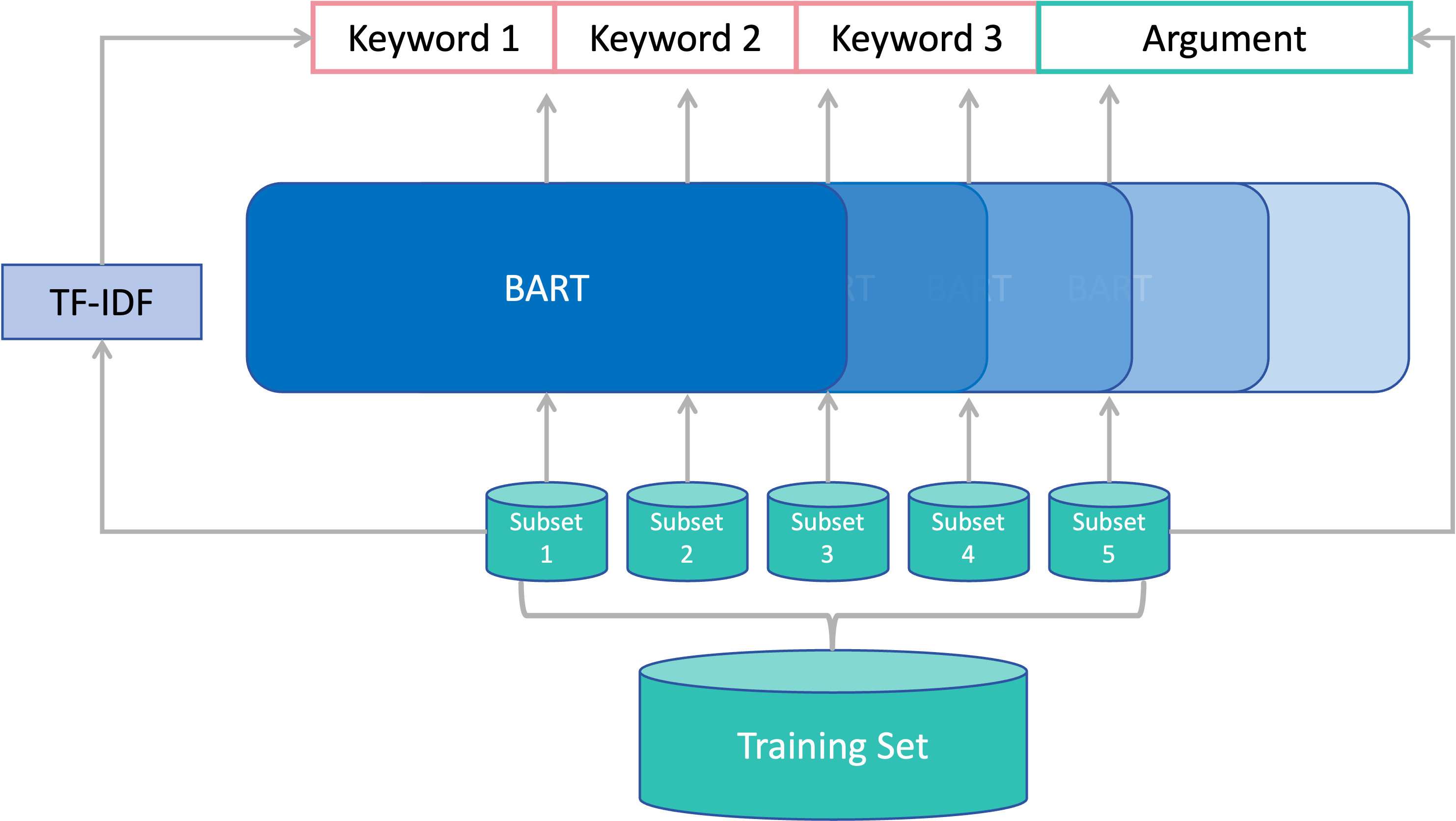}
    \caption{Illustration of our method.}
    \label{track2hitsz1}
\end{figure}

To overcome the identified challenges, a two-part framework (figure~\ref{track2hitsz1}) was devised. 

The first component, Diverse Generation Strategy Based on Subset Division (DiverGS), involved splitting the training data into five exclusive subsets to train five individual BART models, each aimed at generating a single, diverse argument per claim. 

The second component, Generation Enhancement Strategy Based on Keyword Guidance (KeyGuide), introduced keywords for each argument to guide the model during generation. These keywords, extracted using the TF-IDF algorithm, were concatenated to the argument's beginning and served as prompts. This approach resulted in a higher diversity and quality of generated arguments.

\subsubsection{Experiments}
The experimental phase began with data preprocessing, where TF-IDF was used to generate and append three keywords to each argument. The arguments for each claim were evenly divided into five subsets, creating five sub-training sets. The pre-trained models used for the generation task include Mengzi-T5~\cite{zhang2021mengzi}, T5~\cite{raffel2020exploring}, BART~\cite{lewis2020bart}, CPT~\cite{shao2024cpt}, etc. And the bart-base-chinese model was selected for its performance in preliminary experiments. Each subset was then used to fine-tune a separate BART model, resulting in five distinct generation models.

\begin{figure}[h]
    \centering
    \includegraphics[width=0.5\textwidth]{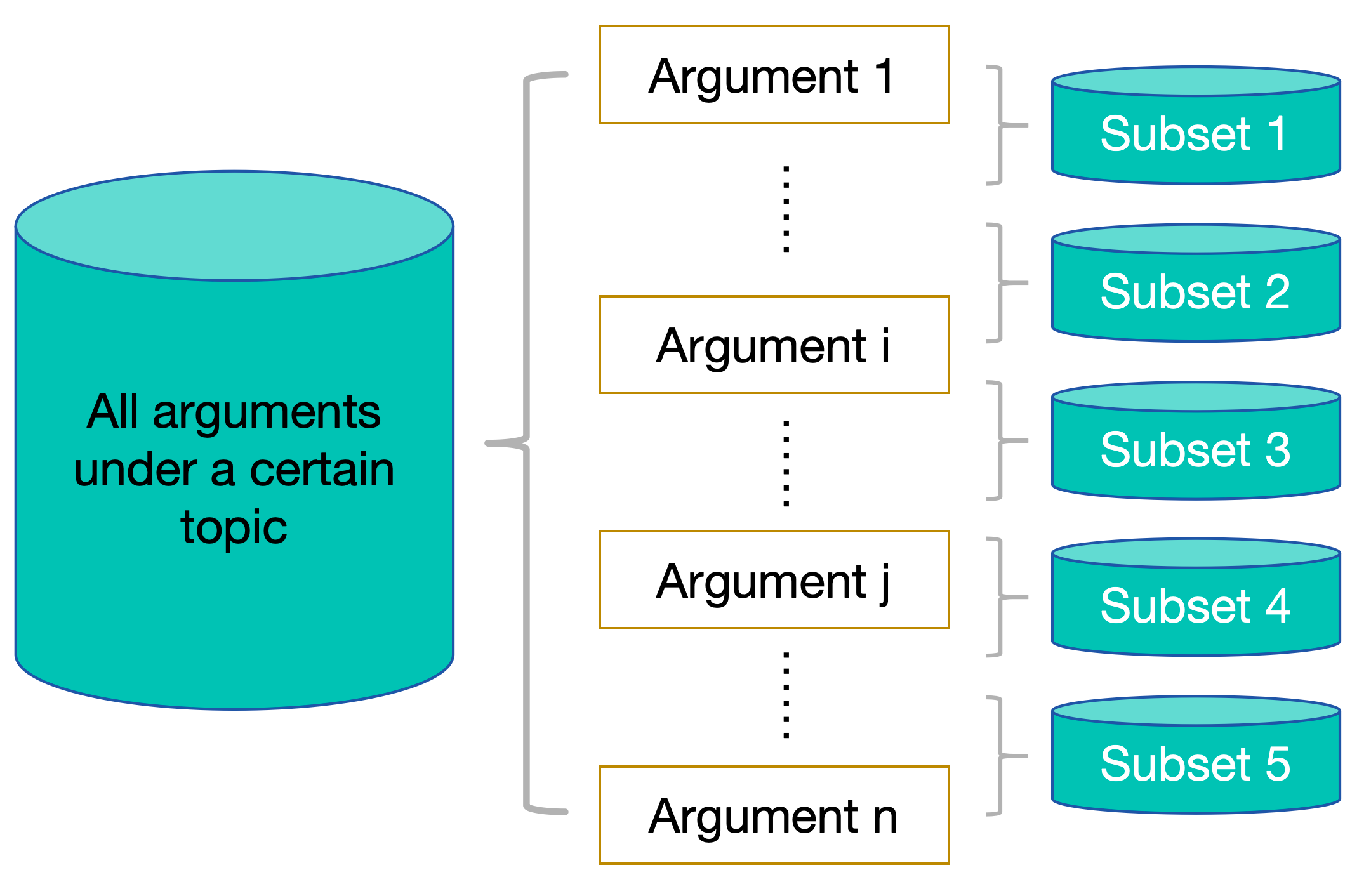}
    \caption{Illustration of subset division.}
    \label{track2hitsz2}
\end{figure}

After obtaining the five subsets, each subset is processed into the form of “source sequence to target sequence” and then used separately to fine-tune five Bart models with different parameters, resulting in five fine-tuned generation models. During the inference phase, the given claims are inputted into the aforementioned five generation models separately, and each model generates one argument. The decoding strategy is beam search, with num\_beams set to 5, maximum sequence length set to 128. Additionally, during decoding, repetition\_penalty is set to 5.0 to alleviate repetition issues, and length\_penalty is set to 5.0.

Our framework achieved a performance of ROUGE-L=0.167 on the official unseen test set provided by the competition committee.

Furthermore, to validate the necessity and effectiveness of each module in our framework, we conducted ablation experiments on the validation set, and the results are shown in Table~\ref{track2hitsz_table1}. It can be seen that compared to not using keywords as guidance, our proposed keyword-guided generation enhancement method leads to a significant improvement in performance. This is because the keywords generated by the model can guide the generation of subsequent arguments. Moreover, our proposed strategy of generating diversity based on subset partitioning shows some improvement in ROUGE-1 and ROUGE-2 scores. This experiment validates the effectiveness of the two modules we proposed.

\begin{table}[h]
\centering
\begin{tabular}{lcccc}
\hline
 & \textbf{BLEU} & \textbf{ROUGE-1} & \textbf{ROUGE-2} & \textbf{ROUGE-L} \\ \hline
Bart & 0.069 & 0.206 & 0.010 & 0.166 \\
Bart w/ KeyGuide & 0.075 & 0.212 & 0.012 & 0.176 \\
Bart-DiverGS & 0.072 & 0.210 & 0.011 & 0.164 \\
Bart-DiverGS w/ KeyGuide & 0.062 & 0.216 & 0.014 & 0.176 \\ \hline
\end{tabular}
\caption{Ablation study results.}
\label{track2hitsz_table1}
\end{table}
\clearpage
\subsection{Track 2: Longest Common Subsequence Search for Claim-based Argument Generation}
This subsection will introduce the details of the model submitted by tingzhidui team in Track 2.

\subsubsection{Solution Description}

The T5 model is used as the baseline model, with the input being "Topic: claim," where the claim is replaced with a specific text.
If the length of the argument is less than 128, other arguments are copied to increase the length. The loss function is the cross-entropy function, with a learning rate of 1e-4 and the optimizer being Adamw. Parameters were adjusted for the T5 and GPT-2 models to achieve the best configuration. After analyzing the experimental results, the GPT-2 model was chosen as the final model.

Our team proposed a method based on two algorithms, a best score~(BS) calculation algorithm based on longest common subsequence~(LCS) and a best standard argument~(BSA) calculation algorithm. Using these algorithms, the best score and the best standard argument for each claim can be obtained.

        
        
          


\begin{figure}[ht]
  \centering
  \setlength{\abovecaptionskip}{-1cm}
  \setlength{\belowcaptionskip}{-1cm}
  \begin{minipage}{0.48\textwidth}
    \includegraphics[height=8cm, width=5.33cm]{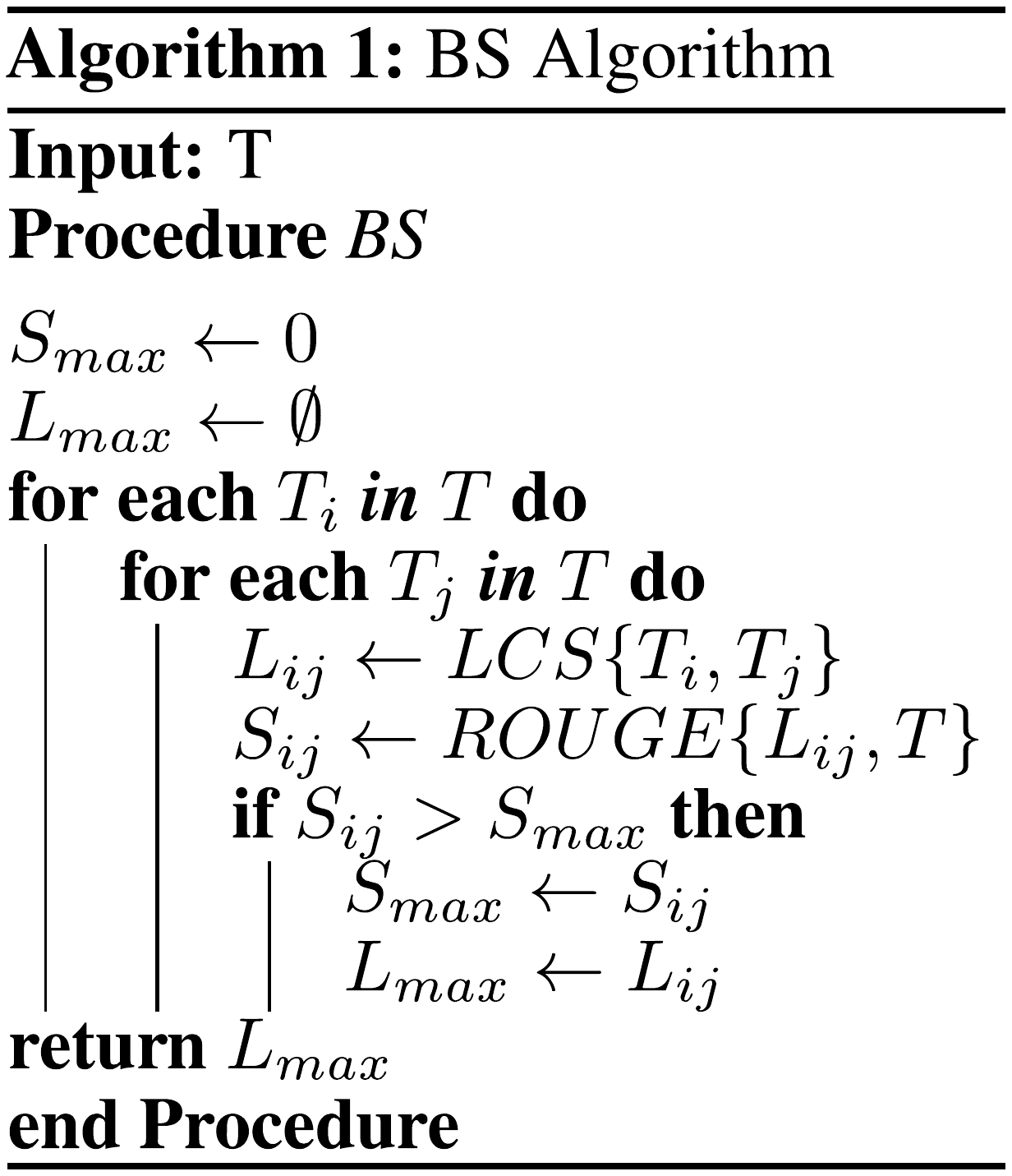}
  \end{minipage}
  \hfill
  \begin{minipage}{0.48\textwidth}
    \includegraphics[height=8cm, width=5.33cm]{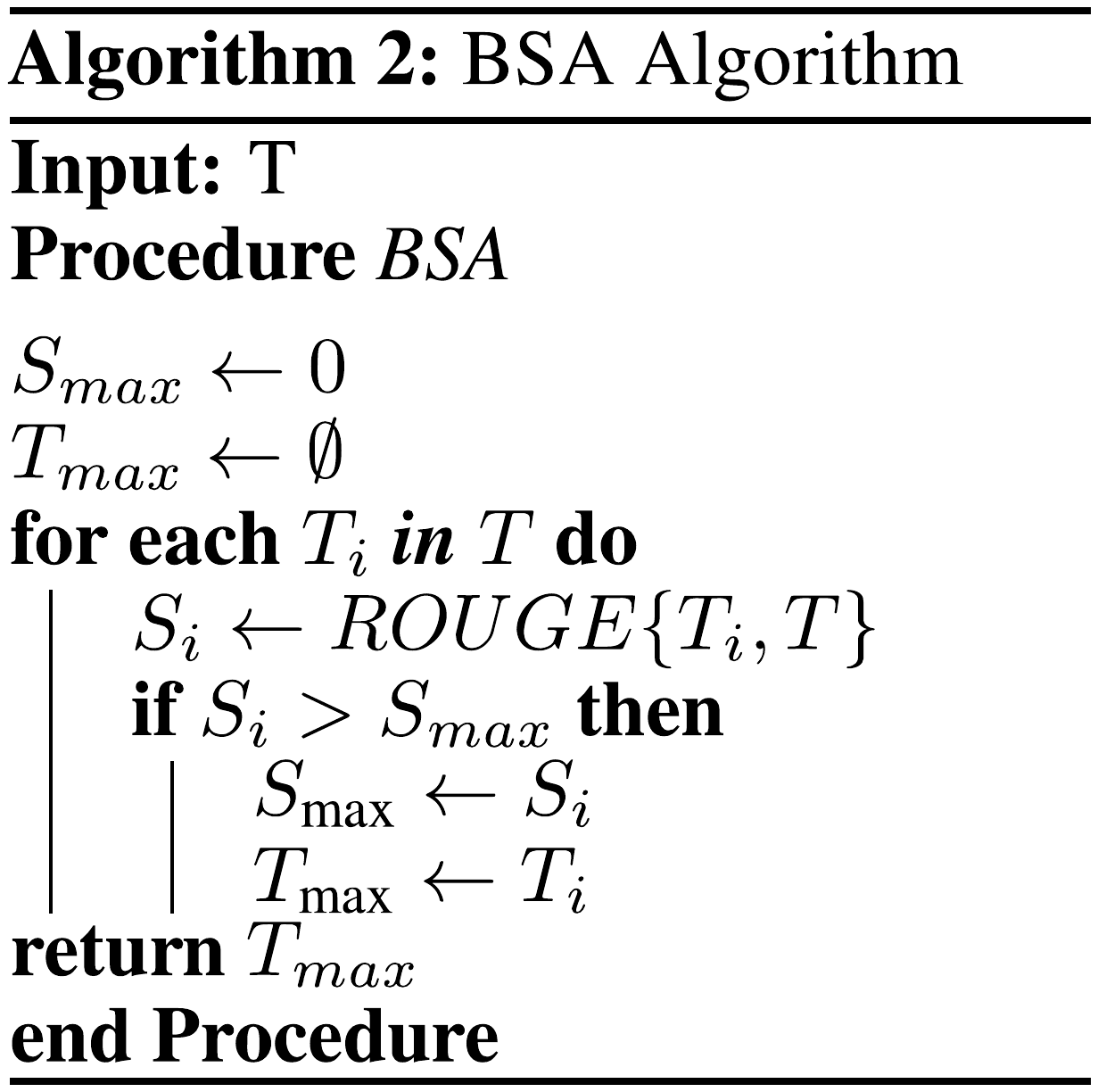}
  \end{minipage}
\end{figure}

\subsubsection{Experiments}
Table~\ref{track2tingzhidui_table1} shows the Rouge-L scores of the T5 and GPT-2 models on the validation set and their performance under different parameter configurations.

Table~\ref{track2tingzhidui_table2} shows some of the best arguments, the scores of the best arguments, the best subsequences, and their scores for certain claims.

\begin{table}[h]
\centering
\begin{tabular}{lcccccc}
\hline
Model & ALC & MCL & MAL & Beam Size & RP & Rouge-L \\
\hline
T5 & T & 32 & 256 & 20 & 5.0 & 0.102 \\
T5 & F & 32 & 256 & 20 & 5.0 & 0.1573 \\
T5 & F & 25 & 256 & 20 & 5.0 & 0.1729 \\
T5 & F & 32 & 128 & 20 & 5.0 & 0.1559 \\
T5 & F & 32 & 256 & 3 & 5.0 & 0.1803 \\
T5 & F & 32 & 256 & 4 & 5.0 & 0.1828 \\
T5 & F & 32 & 256 & 5 & 5.0 & 0.1666 \\
T5 & F & 32 & 256 & 10 & 5.0 & 0.1679 \\
T5 & F & 25 & 256 & 4 & 5.0 & 0.1757 \\
T5 & F & 32 & 256 & 20 & 2.0 & 0.1745 \\
T5 & F & 32 & 256 & 20 & 3.0 & 0.1627 \\
T5 & F & 32 & 256 & 4 & 2.0 & 0.1767 \\
GPT-2 & F & 32 & 256 & 4 & / & 0.1884 \\
\hline
\end{tabular}
\caption{The T5 model and the GPT-2 model's Rouge-L scores on the validation set. We report Argument Length Completion~(ALC), Maximum Claim Length~(MCL), Maximum Argument Length~(MAL), Beam Size during beam search, Repetition Penalty~(RP) and corresponding Rouge-L score.}
\label{track2tingzhidui_table1}
\end{table}

\begin{table}[h]
\centering
\begin{tabular}{p{2cm}|p{6cm}|l||p{2cm}|l}
\hline
Claim & BSA & BSA score & BS & BS Score \\
\hline
“佛系”标签对青年人成长弊大于利 & “首先是在心理学层面，佛系标签不利于青年人成长的人格全面发展。” & 0.2115 & “佛系标签是的，在的不的。” & 0.2449 \\
\hline
“佛系”标签对青年人成长利大于弊 & “综上佛系标签对青年人的成长利大于弊，谢谢。” & 0.2085 & “我们是佛系，是的，的。” & 0.2419 \\
\hline
短视频的火爆是精神文化匮乏的表现 & “而体现二字，一方面指短视频火爆的成因是精神文化的匮乏，另一方面是说短视频本身的精神文化也是匮乏的。” & 0.2294 & “方精神文化的，是短视频的精神文化是的。” & 0.2734 \\
\hline
对知识网红的崇拜让我们对真知更远 & “如果它是提供给你知识可以，可是如果他提供给你的知识方式是告诉你这是你思维的终点，提供知识的同时剥夺你的思维，这不可以。” & 0.2123 & “知识网红是，是的，我。” & 0.2539 \\
\hline
\end{tabular}
\caption{Best Subsequence and Best Standard Argument Results. The claim, best standard argument~(BSA), best subsequence~(BS) and corresponding scores are reported in the table.}
\label{track2tingzhidui_table2}
\end{table}

It was found that the evaluation index did not reflect the differences between predicted arguments and proposed a method based on high-frequency words and keywords. The GPT-2 model was used for training and prediction, achieving a Rouge-L score of 0.2035, and the score after submission was 0.125.

\clearpage
\section{Conclusion}
The AI-Debater 2023 Challenge moves towards argument generation tasks. We set up counter-argument generation and claim-based argument generation tasks. In this challenge, we build and release a new counter-argument generation dataset, enriching argument generation tasks.

The winning approaches, which included data augmentation, instruction tuning, and diffusion model integration, have demonstrated the potential of current AI technologies to understand and construct arguments. These methods have not only improved the performance of the models but also provided insights into how AI can be further developed for complex language tasks.

Looking ahead, the challenge has identified key areas for future research, including enhancing argument quality, addressing data imbalance, and improving coherence in generated texts. As the field progresses, it is expected that AI will increasingly contribute to nuanced debates, offering new possibilities for AI applications in various domains.

\section*{Acknowledgements}
This work is supported by National Natural Science Foundation of China (No. 62176058) and National Key R \& D Program of China (2023YFF1204800). The project's computational resources are supported by CFFF platform of Fudan University.

\clearpage
\bibliographystyle{unsrt}
\bibliography{references}  

\clearpage
\appendix
\section{Challenge Result}
\label{cha_res}
\subsection{Track 1: Counter-Argument Generation}
\begin{table}[h]
\centering
\begin{tabular}{l|l}
\hline
\ \textbf{Team} & \textbf{Score}\\
\hline
\ HITSZ-HLT & 25.2 \\
\hline
\  ZUT & 18.8 \\
\hline
\  huashui & 17.2 \\
\hline
\  tingzhidui & 16.3\\
\hline
\  baseline & 14.3\\
\hline
\end{tabular}
\caption{Performance of participants on Track 1. }
\label{track1res}
\end{table}

\subsection{Track 2: Claim-based Argument Generation}
\begin{table}[h]
\centering
\begin{tabular}{l|l}
\hline
\ \textbf{Team} & \textbf{Score}\\
\hline
\ HITSZ-HLT & 16.7 \\
\hline
\  ZUT & 15.4 \\
\hline
\  tingzhidui & 12.5\\
\hline
\  baseline & 10.1\\
\hline
\end{tabular}
\caption{Performance of participants on Track 2. }
\label{track2res}
\end{table}

\end{CJK}
\end{document}